\documentclass[sn-mathphys,Numbered]{sn-jnl}  % 12pt font required by SPIthatdocumentclainterpret,12pt]{spieman}  % use this instead for A4 paper
\usepackage{amsmath,amsfonts,amssymb}
\usepackage{graphicx}
\usepackage{setspace}
\usepackage{tocloft}
\usepackage{lineno}

\usepackage{color,soul}
\usepackage{multirow}%
\usepackage{amsmath,amssymb,amsfonts}%
\usepackage{amsthm}%
\usepackage{mathrsfs}%
\usepackage[title]{appendix}%
\usepackage{xcolor}%
\usepackage{textcomp}%
\usepackage{manyfoot}%
\usepackage{booktabs}%
\usepackage{algorithm}%
\usepackage{algorithmicx}%
\usepackage{algpseudocode}%
\usepackage{listings}%

\theoremstyle{thmstyleone}%
\theoremstyle{thmstyletwo}%

\theoremstyle{thmstylethree}%

\begin{document} 
%\title{Intepretative Deep Learning via Domain Adaptation for Fluorescence Spectroscopy}
%\title{Physico-chemical latent modelling via deep learning and Domain Adapation for Fluorescence Spectroscopy}
\title{Deep Learning Domain Adaptation to Understand Physico-Chemical Processes from Fluorescence Spectroscopy Small Datasets: Application to Ageing of Olive Oil}

\author[1,2]{\fnm{Umberto} \sur{Michelucci}}\email{umberto.michelucci@toelt.ai}

\author*[1,3]{\fnm{Francesca} \sur{Venturini}}\email{francesca.venturini@zhaw.ch}

\affil[1]{\orgdiv{Research and Development}, \orgname{TOELT LLC}, \orgaddress{ \city{Dübendorf}, \postcode{8600}, \country{Switzerland}}}

\affil[2]{\orgdiv{Computer Science Department}, \orgname{Lucerne University of Applied Sciences and Arts}, \orgaddress{\city{Rotkreuz}, \postcode{6343},  \country{Switzerland}}}

\affil*[3]{\orgdiv{Institute of Applied Mathematics and Physics}, \orgname{Zurich University of Applied Sciences}, \orgaddress{\city{Winterthur}, \postcode{8401}, \country{Switzerland}}}

\renewcommand{\cftdotsep}{\cftnodots}
\cftpagenumbersoff{figure}
\cftpagenumbersoff{table} 

\abstract{
Fluorescence spectroscopy is a fundamental tool in life sciences and chemistry, widely used for applications such as environmental monitoring, food quality control, and biomedical diagnostics. However, analysis of spectroscopic data with deep learning, in particular of fluorescence excitation-emission matrices (EEMs), presents significant challenges due to the typically small and sparse datasets available. Furthermore, the analysis of EEMs is difficult due to their high dimensionality and overlapping spectral features. This study proposes a new approach that exploits domain adaptation with pretrained vision models, alongside a novel interpretability algorithm to address these challenges. Thanks to specialised feature engineering of the neural networks described in this work, we are now able to provide deeper insights into the physico-chemical processes underlying the data. The proposed approach is demonstrated through the analysis of the oxidation process in extra virgin olive oil (EVOO) during ageing, showing its effectiveness in predicting quality indicators and identifying the spectral bands, and thus the molecules involved in the process. This work describes a significantly innovative approach in the use of deep learning for spectroscopy, transforming it from a black box into a tool for understanding complex biological and chemical processes.}
% Fluorescence spectroscopy, crucial in life sciences, faces challenges in integrating deep learning due to data scarcity. This study leverages transfer learning and domain adaptation with pretrained vision models coupled with a newly proposed interpretability approach to enhance data analysis. We present a method for improving feature recognition in large networks and gain deeper insights into the underlying physico-chemical processes, laying the groundwork for advanced interpretive models in life sciences.

% Include a list of up to six keywords after the abstract

\maketitle
%\linenumbers

\begin{spacing}{1}   % use double spacing for rest of manuscript  % use double spacing for rest of manuscript

%\section{Introduction}

Fluorescence spectroscopy is a central analysis tool in life sciences and chemistry, with applications ranging from environmental monitoring, food quality control, to biomedical diagnostics. Furthermore, it is employed at all scales, from single molecules to tissues and organs, from protein dynamics to in-vivo imaging due to its sensitivity and specificity \cite{moerner2003methods,lakowicz2006principles, dos2022alzheimer}. 
Fluorescence excitation-emission matrices (EEMs), in particular, provide detailed insights into the absorption and emission characteristics of substances, thereby acting as an effective fingerprinting tool \cite{sikorska2019fluorescence,costa2020identification}. However, analysis of EEM data presents significant challenges because of its high dimensionality and the frequent presence of overlapping spectral features. These challenges are traditionally addressed with multivariate chemometric methods such as principal component analysis (PCA) and parallel factor analysis (PARAFAC) \cite{bro1997parafac}. 

Deep learning (DL) has shown the ability to enhance scientific insight through advanced pattern recognition across multiple disciplines \cite{litjens2017survey}. However, the effectiveness of hidden layers, critical for feature extraction and pattern decoding, is highly dependent on having access to extensive and varied datasets \cite{Michelucci2023AppliedDL}. Thus, despite its potential, DL is still not widely used to explain and interpret biological, physical, and chemical processes from spectroscopic data because of (i) methodological and (ii) data-related challenges. Methodologically, DL architectures are designed primarily for computer vision tasks, focus on model decision making rather than understanding the underlying processes, and do not account for the correlation between signals at different wavelengths, treating them as independent features \cite{meza2021applications,liu2021survey}. Data-related challenges arise because spectroscopy datasets are often small or sparse, exhibit either high similarity among images or excessive variability, and are frequently unbalanced \cite{yu2015impact,lahnemann2020eleven}. 
%\begin{itemize}
%    \item[(i)] 
%DL architectures are designed for computer vision tasks, focus on model decision making rather than understanding the underlying processes, and do not account for the correlation between signals at different wavelengths, treating them as independent features \cite{meza2021applications,liu2021survey}.
%\item[(ii)]
%Life science datasets are often small or sparse, exhibit either high similarity among images or excessive variability, and are frequently unbalanced \cite{yu2015impact,lahnemann2020eleven}. 
%\end{itemize}
These challenges hinder the training of large neural networks (NNs), resulting in high variance and inadequate performance, thus undermining confidence in the model and its feature engineering capabilities. Additionally, the instability of NNs prevents effective linking of feature extraction in hidden layers to the phenomena being modelled, as the networks may overfit to noise rather than learning meaningful patterns from the data. 

Here we describe a novel method for spectroscopy applications that works for small and sparse datasets and that transforms DL into a tool to understand biological or physico-chemical processes. The method we propose is based on domain adaptation and a novel intepretability approach. 
This is demonstrates by analysing the natural oxidation process that occurs in extra virgin olive oil (EVOO) during storage, which deteriorates its quality \cite{venturini2024shedding} and reduces its beneficial impact against the risk of cardiovascular and all-cause mortality \cite{donat2023only}. The quality of EVOO is assessed through a series of parameters defined by the United Nations Food and Agriculture Organisation and the European Union \cite{regulation1991commission, european2013commission,FAO}. Among these, the quality indicators $K_{232}$ and $K_{268}$, corresponding to the UV absorbance at 232 and 268 nm, respectively, were chosen for this study as they measure primary and secondary oxidation products.

This work presents two major contributions. The first contribution is the proposed domain adapation approach to design and train a DL model that performs a regression task with exceptional performance even on a limited dataset. The approach is applied to the prediction of the $K_{232}$ and $K_{268}$ quality indicators from fluorescence EEMs.
The second is the description of a method to extract the learned internal representation from the trained DL model without any a priori knowledge or feature engineering. The approach then determines the most relevant spectral bands related to the process and, as such, indicates the chemical components in EVOO involved in the oxidation process.

\section*{Material and Methods}

\begin{figure}[hbt]
    \centering
    \includegraphics[width=0.8\textwidth]{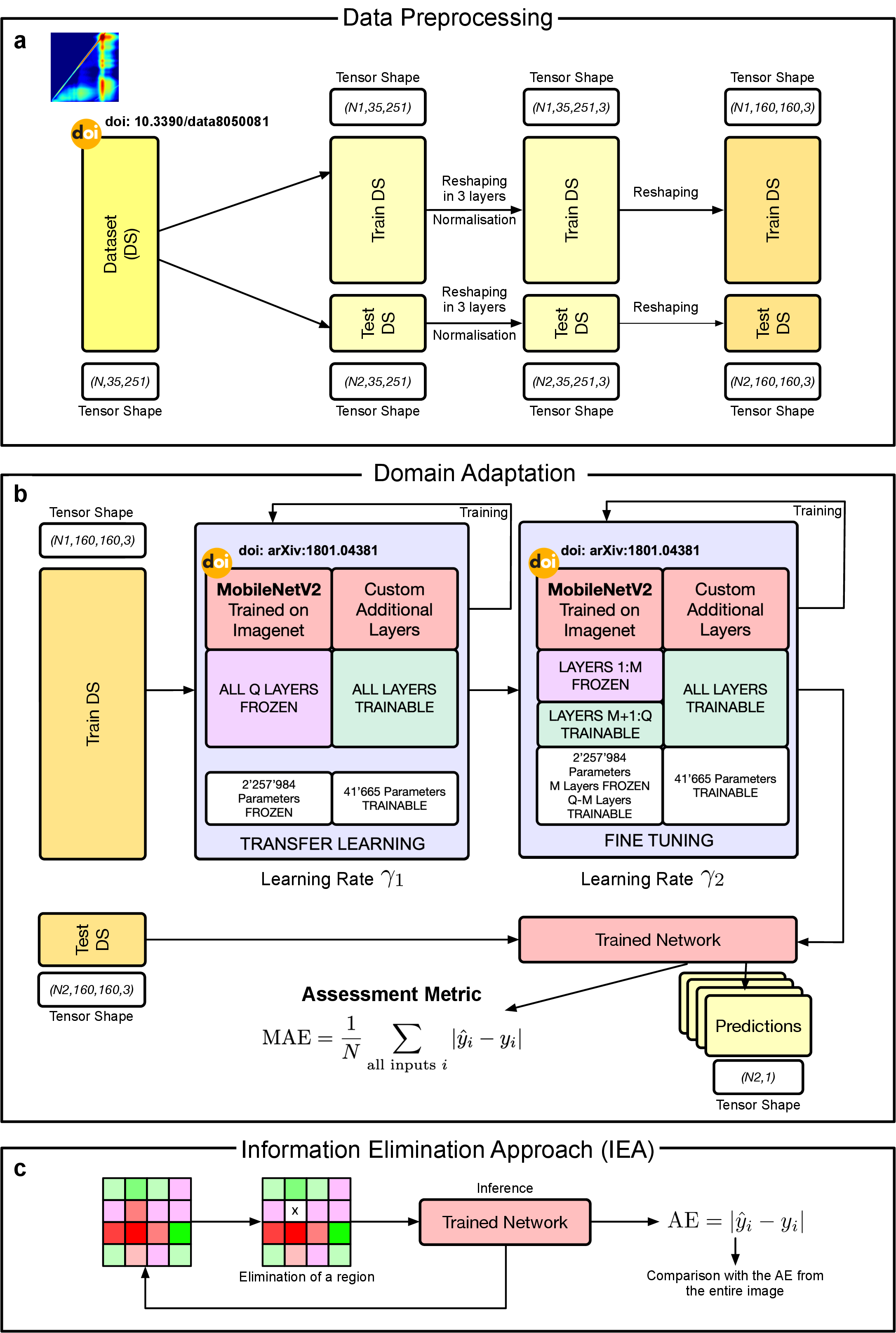}
    \caption{Overview of the  phases of the machine learning approach. \textbf{a}, The data preprocessing phase consists of splitting the data set for the LOO approach, normalisation of pixel values, and preparation for the MobileNetv2 network input layer by reshaping and creating the three necessary layers. \textbf{b}, The transfer-learning and fine-tuning phases allow the network to learn relevant features and create an internal representation of the physico-chemical models that can then be used for the interpretation. \textbf{c}, Information Elimination Approach (IEA) process diagram. \textbf{doi} references indicate the papers that describe some of the used components.}
    \label{fig:transferlearning-finetuning}
\end{figure}

The proposed approach is divided into three steps, shown schematically in Figure \ref{fig:transferlearning-finetuning}: data preprocessing, domain adaptation, and the information extraction approach. Domain adaptation (a technique where a model trained on data from one domain is adapted to perform well in another domain) improves the feature engineering capacity of neural networks \cite{ghafoorian2017transfer}, but is, to the best knowledge of the authors, never been implemented in spectroscopy. 
The challenge of a small and sparse dataset is adressed by employing domain adapation based on the network MobileNetv2 \cite{sandler2018mobilenetv2} (154 layers and ca. 3.5 million parameters) that has been pretrained on the ImageNet dataset \cite{5206848} (containing more than 14 million images and 20000 classes) and is known for its feature extraction capabilities. 
%The first necessary step is the preprocessing of the data, as the pretrained network MobileNetv2 expects data to be in a specific format (Figure \ref{fig:transferlearning-finetuning}a). 
Domain adaptation is composed of two phases: transfer learning and fine tuning (Figure \ref{fig:transferlearning-finetuning}b). This last step addresses the deficiency that the model has never seen any EEMs in its training data.
% For the transfer-learning phase, we keep the feature extraction layers (the backbone of MobileNetv2) frozen and add to the NN one global averaging and three dense layers that learn to predict specific process-related quantities. 
% To make the network learn what features are relevant to the process of interest, we proceeded with the second phase, the fine-tuning. Here, we unfreeze the last 54 layers of the MobilNetv2 model and continue the training (Figure \ref{fig:transferlearning-finetuning}b). 
Finally, to investigate the oxidation process, we introduce the Information Elimination Approach (IEA) (Figure \ref{fig:transferlearning-finetuning}c). The IEA is the key to extract information from the experimental observations. The method consists in removing part of the spectral information feeded to the network as input, and evaluating the model performance drop, thus assessing the importance of the removed information by its impact on the predictions.

\subsection*{Fluorescence Excitation-Emission Matrices and UV Absorption Dataset}

The dataset comprises fluorescence spectroscopy excitation-emission matrices (EEMs) and UV absorption spectroscopy data for 24 commercial extra virgin olive oils (EVOO), fresh, and in 9 oxidation stages. The oils were chosen to be as heterogeneous as possible in both the origin of production and the price to identify general features of the oxidation process. The oils underwent accelerated oxidation at 60 $^\circ$C to investigate the impact of long-term storage. The effects of thermal degradation were evaluated at ten intervals over a time of 53 days. Fluorescence emission and UV absorption measurements at each stage provided insights into the oxidation process, which is responsible for changes in the oil's chemical properties.
The raw data for the fluorescence spectroscopy EEMs, acquired with an Agilent Cary Eclipse Fluorescence Spectrometer, are stored in CSV files, detailing fluorescence intensities across various excitation and emission wavelengths. Each file contains the fluorescence emission from 300 nm to 800 nm in 2 nm increments (251 values), under excitation at 35 wavelengths ranging from 300 nm to 640 nm in 10 nm increments. All data were acquired under identical conditions and at a controlled temperature of 22 $^\circ$C within a thermalized sample holder, so the intensities are directly comparable.
%Notably, the dataset includes intensities from Rayleigh scattering, evident when the emission wavelength matches the excitation wavelength; these values have been retained in the raw data. Their influence on the results described in this paper was investigated by removing and substituting the data either with interpolated or 0 values. No changes in the finding was observed. 
For UV spectroscopy measurements, the olive oil samples were diluted in isooctane and prepared according to EU regulations \cite{regulation1991commission, european2013commission}. The analysis was carried out in sealed quartz cuvettes under identical conditions with an Agilent Cary 300 UV–Vis spectrophotometer and at a controlled temperature of 22 $^\circ$C within a thermalized sample holder, collecting the extinction coefficients at four specific wavelengths: 232 nm, 264 nm, 268 nm and 272 nm.

The complete dataset is available for download, and is described in detail in \cite{venturini2023dataset} where the link to download the data can be found.

\subsection*{Domain Adaptation}

Domain adaptation, in the context of machine learning, is a technique that involves adapting a model trained on data from one domain to perform well on data from another domain (hence the name). This kind of approach is widely used for improving the model's generalisation ability and performance on new, unseen data that differ from the model's training data.
%\subsection*{Pre-processing for Domain Adapation}
To apply this approach in the case described in this work, it is necessary to prepare the data according to a set of steps. Firstly, the dataset was split for leave-one-out (LOO) cross-validation to ensure that each sample was validated exactly once. That means that for each oil the model was trained on 23 oils and 10 oxidation stages, and validated on the left out oil and its 10 oxidation stages. This is repeated for all oils. EEM's pixel values (fluorescence intensities) ranged from 0 to 1000 counts, therefore, the intensity values were normalised in the dataset by dividing pixel values by a fixed value of 1000 to ensure no data leakage. Furthermore, data was reshaped to comply with the input requirements of the MobileNetv2 network \cite{sandler2018mobilenetv2}, involving adjustments to \(160 \times 160\) pixels, reformatting to three channels, and conversion to an unsigned 8-bit integer format. The three channels were synthetically generated by triplicating the measured fluorescence intensity (so all three channels are identical), as depicted in Figure \ref{fig:transferlearning-finetuning}a. Intensities from Rayleigh scattering, where the emission wavelength matches the excitation wavelength, were retained in the raw data. These intensities are typically considered not relevant, but we wanted to check if the network ignored them or not (the network, in the correct way, ignores the Rayleigh scattering). We also tested the results by removing the Rayleigh scattering (either by setting pixel values to 0 around it or by interpolating spectra around it) and we observed no difference in the results.
These preprocessing steps are the key to adequately preparing the data for effective feature extraction and subsequent analysis using the pretrained MobileNetv2 network.

%\subsection*{Domain Adaptation (Model Training)}

The model training is then performed according to the following phases, depicted in Figure \ref{fig:transferlearning-finetuning}b.
\begin{enumerate}
    \item \textbf{Phase I - Transfer learning}: in a first phase a network is built and trained according to the following recipe:
    \begin{itemize}
        \item Start with the MobileNetv2 \cite{sandler2018mobilenetv2} network with the weights obtained by training it with the imagenet dataset.
        \item Remove from the MobileNetv2 all the dense layers after the backbone.
        \item Freeze the backbone of MobileNetv2. The backbone will not be trained during this phase.
        \item Add the following sequence of layers to the MobileNetv2 backbone: (i) global averaging, (ii) a dropout layer with a factor of 0.2, (iii) a dense layer with 32 neurons with a ReLU activation function, (iv) a dense layer with 16 neurons with a ReLU activation function, (v) a dense layer with 8 neurons with a ReLU activation function,
        (vi) a dense layer with 1 neuron with an identity activation function.
        \item Train the network with the Adam optimiser. For all parameters $K_{232}$ and $K_{268}$ we used the following parameters: learning rate $\gamma=10^{-4}$, mean squared error (MSE) as loss function, mini-bact size $b=230$, and 1000 epochs. 
    \end{itemize}
    \item \textbf{Phase II - Fine-tuning}: in a second phase the  training proceeds according to the following steps.
    \begin{itemize}
        \item Unfreeze the last 54 layers of the MobilenNtv2 backbone. During the training process, the initial 100 layers of the network remain frozen, with subsequent layers being actively trained. The decision on the number of layers to unfreeze was based on multiple tests, which indicated minimal variation in performance when the number of frozen layers ranged from 100 to 120.
        \item Train the network with the Adam optimiser. For the parameters $K_{232}$ we used the following parameters: learning rate $\gamma=10^{-6}$, mean squared error (MSE) as loss function, mini-batch size $b=230$, for 500 epochs. For the parameters $K_{268}$ we used the following parameters: learning rate $\gamma=10^{-5}$, MSE as loss function, mini-batch size $b=32$, and 500 epochs. 
    \end{itemize}
\end{enumerate}
%The dense layers are trained on 230 EEMs using, as mentioned, a leave-one-out cross-validation approach.
The data set comprises 24 oils and thus has a small size. To perform cross-validation, we chose the leave-one-out (LOO) approach. LOO cross-validation works by using all observations from the original sample except one (23 oils at their oxidation stages) as training data and the left-out single observation (one oil at its 10 oxidation stages) as validation data.
This process is repeated so that each oil in the sample is used once as validation data. This method is particularly useful for our small dataset because it maximally uses the data for training while still ensuring that each data point is validated. The dataset is partitioned on the basis of individual oils to prevent data leakage, ensuring that all oxidation stages for a specific oil are grouped together in the same split.

\subsection*{Information Elimination Algorithm}

The Information Elimination Algorithm (IEA), presented in this paper, exploits the internal feature representation of large NNs to explain and understand natural processes. It is crucial to emphasise that our aim is not to comprehend the thought process of the model, but rather to utilise the acquired knowledge to gain insights into the oxidation process. This algorithm systematically removes specific regions from the EEMs and observes the impact on the model's performance, particularly focussing on the prediction errors.
Specifically, the algorithm works as follows.
Consider the trained model that has been validated on oil $j$. 
We proceed by first removing a region of \(5\times5\) pixels from the EEM of oil \(j\) and assessing the impact on the prediction of \(K_{232}\) and $K_{268}$ by recording the absolute error (AE).
The process begins at the top left corner and proceeds horizontally towards the right edge. Upon reaching the end of a row, it continues from the leftmost side of the row immediately below, repeating this pattern until the entire image have been covered. 
This process is repeated iteratively until the importance of all portions of the image to the model's decision-making process is assessed. 
The result is a heatmap that identifies the spectral bands relevant to the prediction and indicative of the physico-chemical oxidation process. For clarity, the heatmap is smoothed with a Gaussian filter (\(\sigma = 3\) pixels), and contour lines are added to illustrate how AE increases are distributed. This technique leverages the network internal feature representation to study natural processes without retraining the model. 

This method allows us to identify and visualize which parts of the EEMs are most critical for accurate predictions, and thus, as a proxy, to the physico-chemical process of oil oxidation, enhancing the interpretability of the DL model. This approach essentially maps the sensitivity of the model's predictions to variations in different spectral areas of the data. This model has been inspired by existing methods, such as the backward feature elimination and $Y$-randomisation approach (see, for example, \cite{rucker2007randomization}), although it uses a fundamentally different approach. To understand the physico-chemical processes underlying the observations, we selectively remove bands associated with different substances and processes, such as chlorophyll (that has a fluorescence emission in the 600 nm - 750 nm for example) or primary and secondary oxidation products (that have a fluorescence emission in the 375 nm - 550 nm range), from the EEMs to evaluate their importance for the predictions. Assuming that the NN has developed an internal representation that models these physico-chemical processes, we can deduce which substances or processes are most influential in the oxidation of the oils.

The size of the removed region (5 $\times$ 5 pixels) corresponds to the increment in the excitation wavelengths of the original data (before reshaping). In the original EEMs (35 $\times$ 251 pixels), the sampling step for the excitation wavelength is 10 nm. 5 pixels in the reshaped matrix (160 $\times$ 160 pixels), corresponds to 11 nm (1 pixel corresponds to 2.2 nm). In the other dimension (emission wavelengths) one could also a take smaller size. However, because of the broad fluorescence features in the EEMs, this will not lead to better intepretability. Thus, we decided to choose a square shape for the region to be removed.

\section*{Results and Discussion}

\begin{figure}[hbt]
    \centering
    \includegraphics[width=1\textwidth]{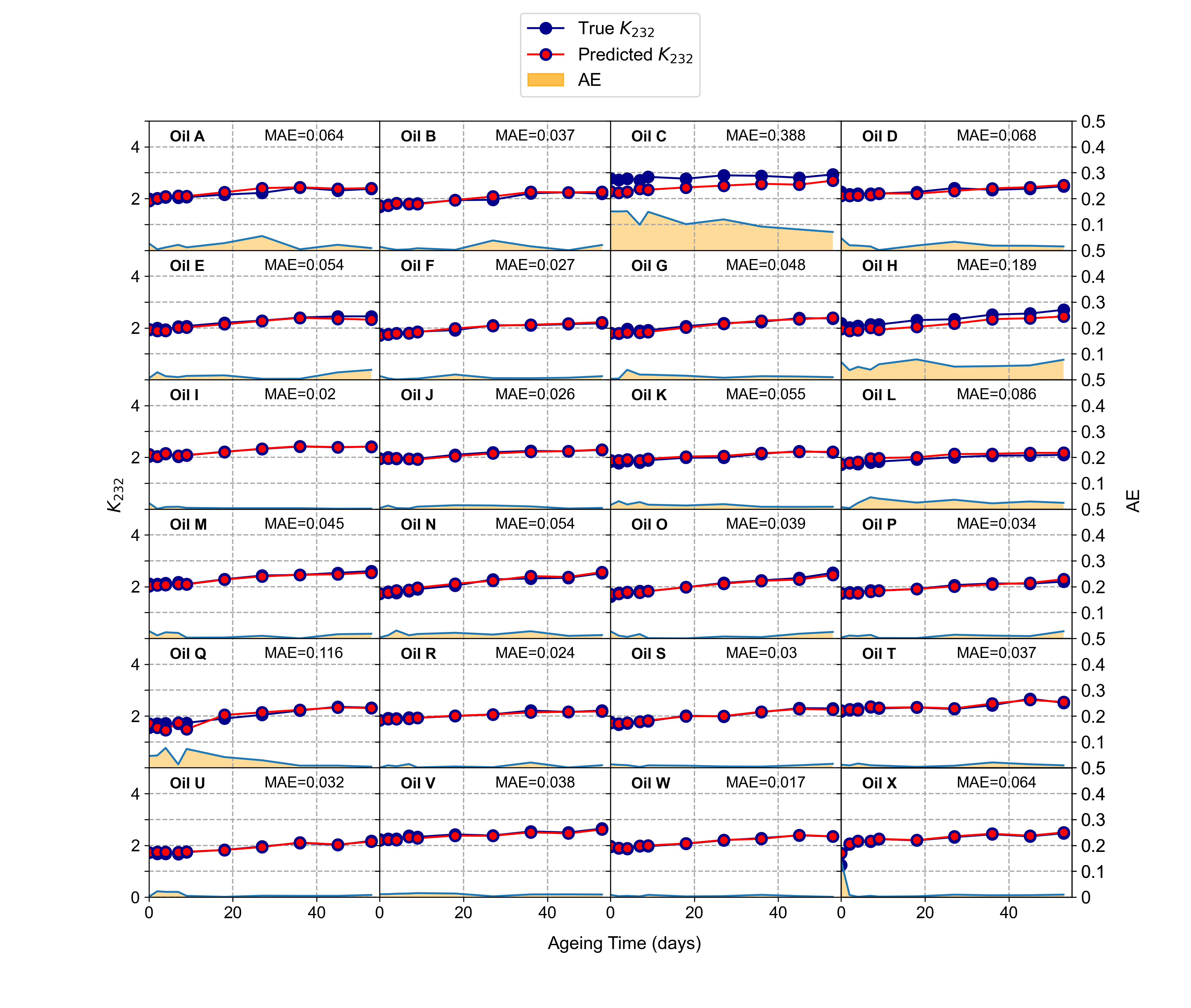}
    \caption{Comparison of the true (blue) and predicted (red) values of the quality indicator $K_{232}$ for all the oils at all oxidation stages (vertical scale on the left axis). The corresponding absolute error (AE) is shown as an area in yellow (vertical scale on the right axis). The Mean Absolute Error (MAE) obtained as average over all the oxidation stage is displayed in each panel for each oil.}
    \label{fig:grid1}
\end{figure}

\begin{figure}[hbt]
    \centering
    \includegraphics[width=1\textwidth]{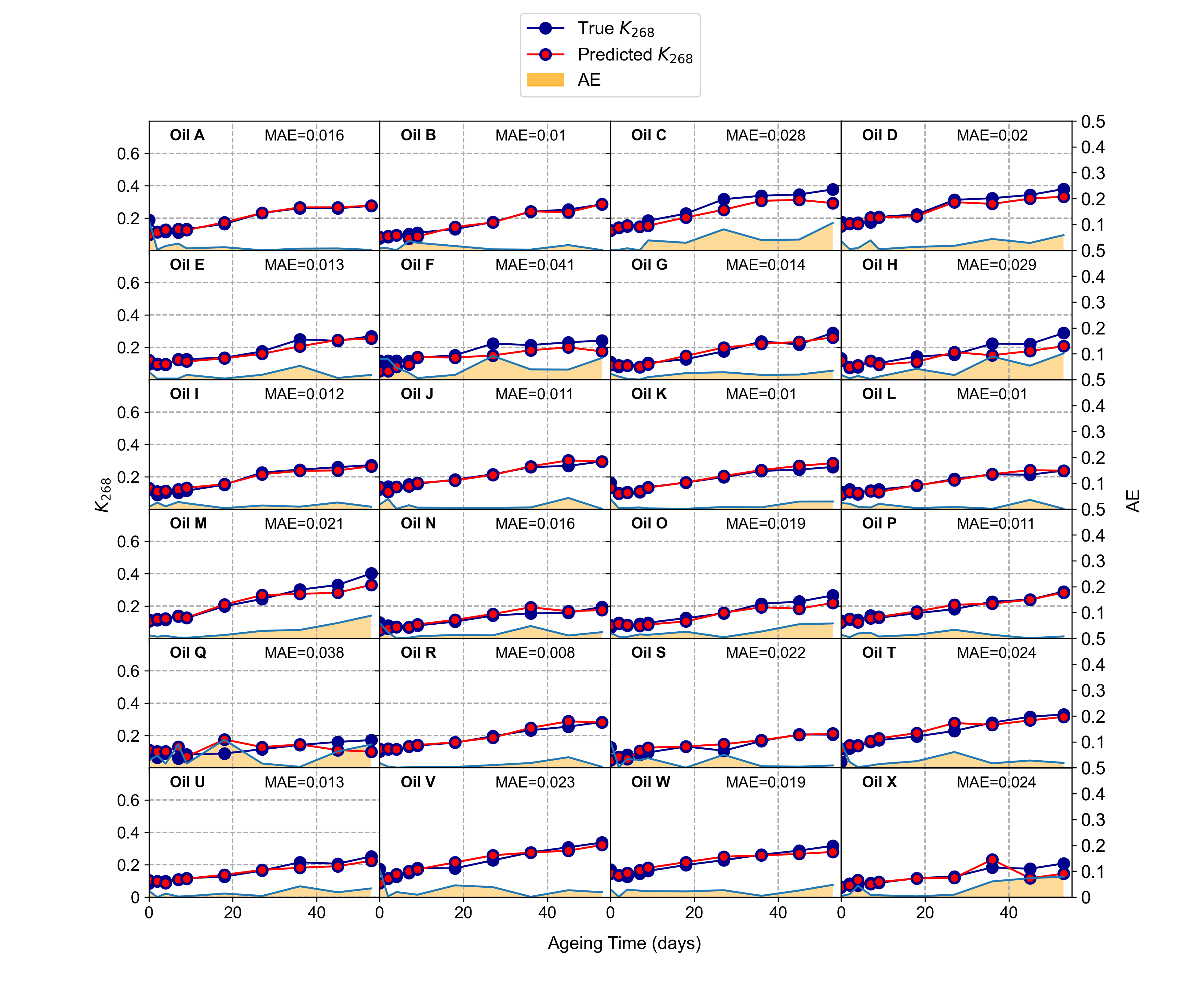}
    \caption{Comparison of the true (blue) and predicted (red) values of the quality indicator $K_{268}$ for all the oils at all oxidation stage (vertical scale on the left axis). The corresponding absolute error (AE) is shown as an area in yellow (vertical scale on the right axis). The Mean Absolute Error (MAE) obtained as average over all the oxidation stage is displayed in each panel for each oil.}
    \label{fig:grid2}
\end{figure}

The exceptional performance of the model in the prediction of the quality indicators $K_{232}$ and $K_{268}$ from fluorescence EEMs for each oil in each oxidation stage is shown in Figures \ref{fig:grid1} and \ref{fig:grid2}, where the predicted and measured values are compared directly.

\begin{figure}[hbt]
    \centering
    \includegraphics[width=0.75\textwidth]{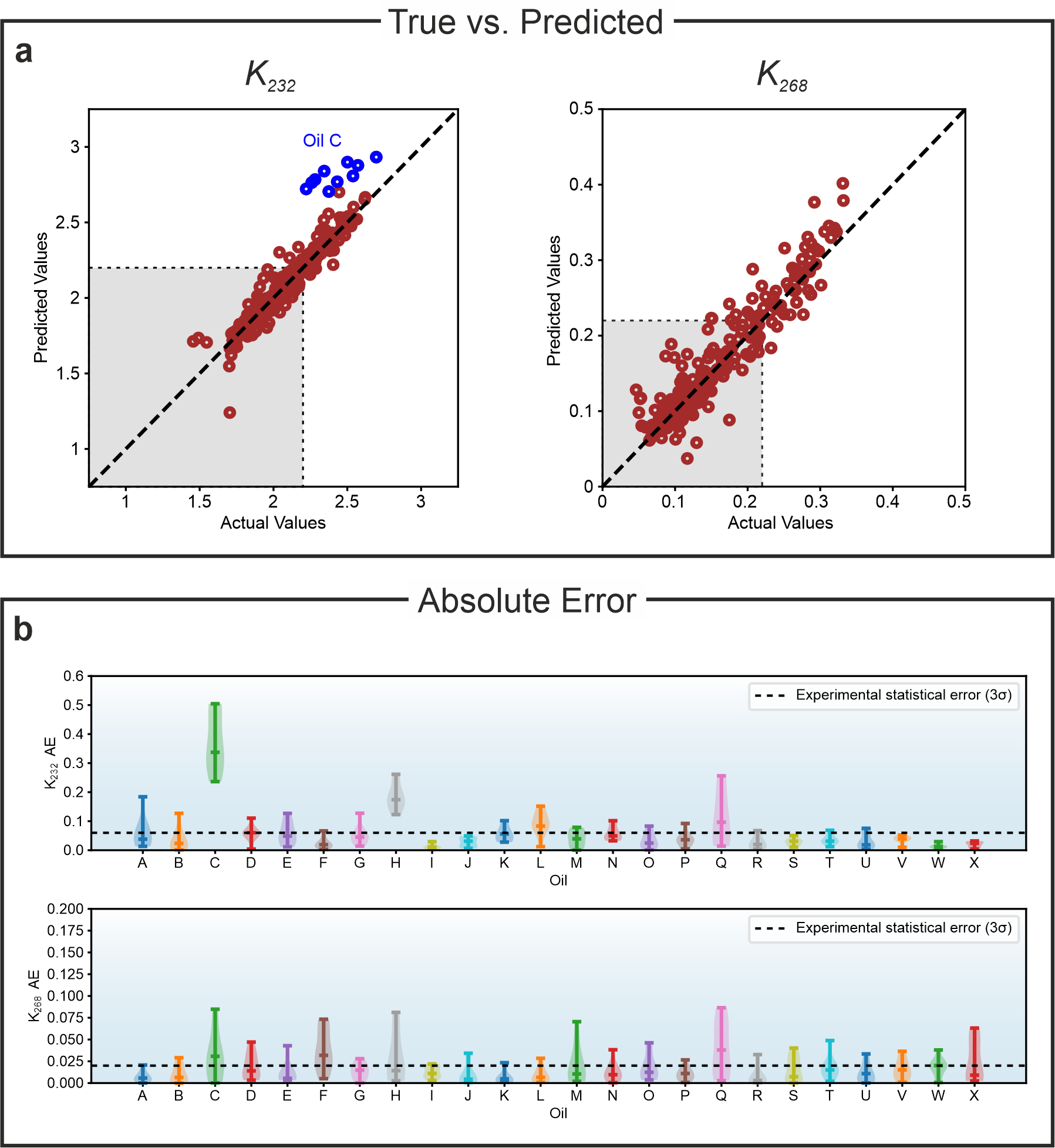}
    \caption{\textbf{a}, Comparison of predicted and measured (actual) values of the quality indicators $K_{232}$ and $K_{268}$ for all oils at all oxidation stages. The gray area in each plot marks the limit set by the Food and Agriculture Organisation of the United Nations and by the European Union. Oil C is marked in blue as the $K_{232}$ value was was already above this limit at the beginning of the study and, therefore, is not well predicted by the model. \textbf{b}, Violin plots of the AE for each oil for $K_{232}$ (above) and $K_{268}$ (below). The dashed lines indicate the 3$\sigma$ statistically estimated experimental error.}
    \label{fig:results1}
\end{figure}

The results are summarised in Figure \ref{fig:results1}. The exception is the prediction for oil C, marked in blue in Figure \ref{fig:results1}a, which is, however, easily explained, as the value $K_{232}$ at the beginning of the study was already above this limit and, therefore, is cannot be well predicted by the model.
Figure \ref{fig:results1}b shows the distribution of the absolute error (AE) for each oil at each oxidation stage. The total mean absolute error (MAE) is 0.066 for $K_{232}$ and 0.010 for $K_{268}$, compatible or lower than the statistically estimated experimental error, also shown in the figure. 

\begin{figure}[hbt]
    \centering
    \includegraphics[width=1.0\textwidth]{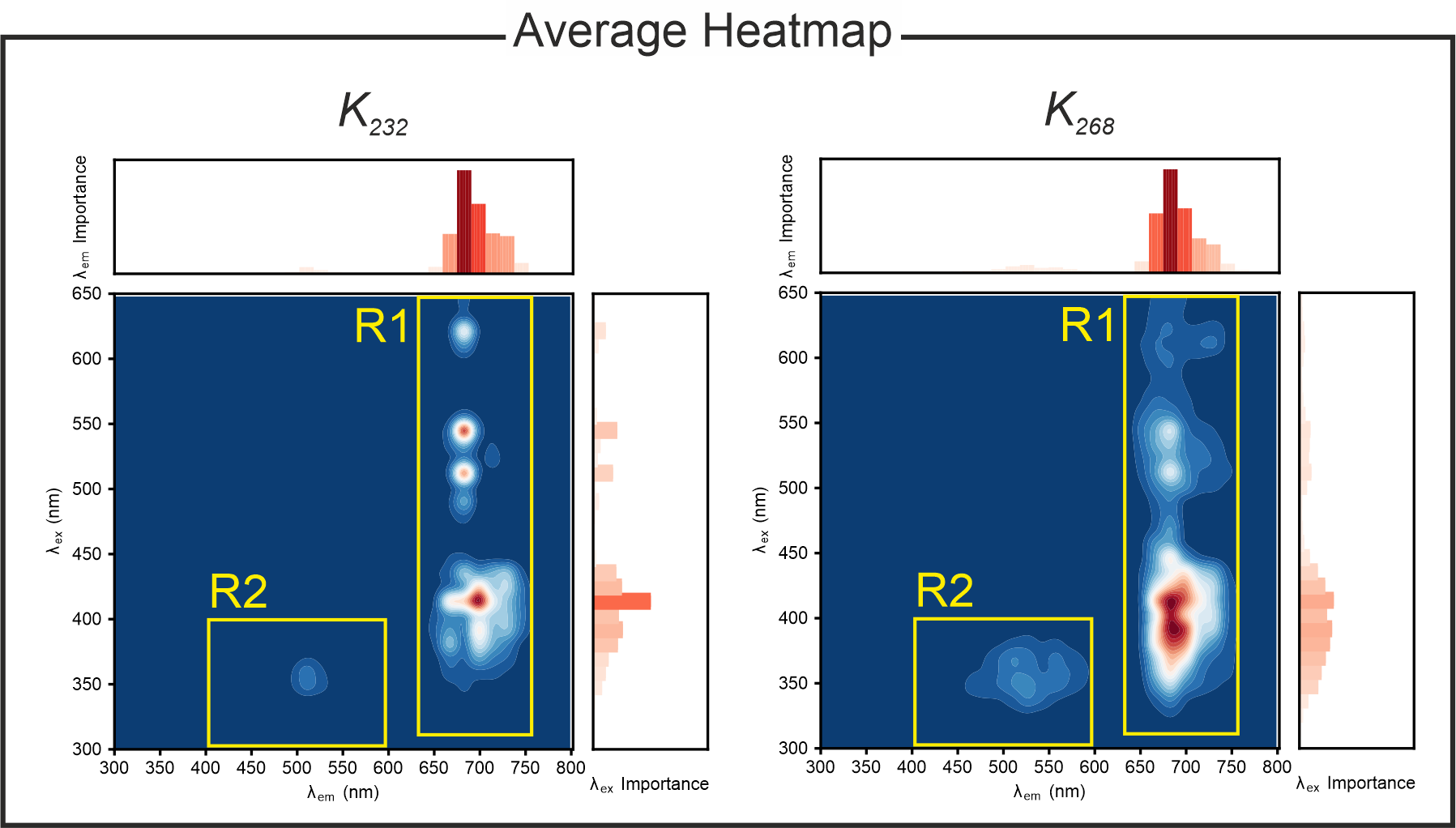}
    \caption{Average of the heatmaps obtained for all oils in the last oxidation stage showing the spectral band of relevance for the prediction of the $K_{232}$ and $K_{268}$. R1 marks the absorption and emission bands of chlorophylls, R2 those of oxidation products.}
    \label{fig:results2}
\end{figure}

Once the reliability of the model has been established, the IEA allows to interpret how the model extracts knowledge from the spectral data. To determine the features that are generally relevant for all EVOO independently of the specific geographical origin or cultivar, the average of the resulting heatmap can be calculated. The result is shown in Figure \ref{fig:results2} for both quality indicators $K_{232}$ and $K_{268}$.
%The bars in the top and right panels have the following meaning.
The bars in the top barplot are 5 pixel wide and have a height corresponding to the sum of all the increases of the AE in the 5-pixel-wide vertical region of the heatmap that lies immediately below the bar. Each bar in the right barplot is analogously 5 pixel wide and has a height corresponding to the sum of all the increases of the AE in the 5-pixel wide vertical region of the heatmap that lies immediately on the left of the bar.
As such, each bar measures the importance of a specific excitation (for the side barplot) and emission (for the top barplot) wavelength range for the predictions of the two parameters $K_{232}$ and $K_{268}$ respectively.

The IEA identifies two relevant spectral ranges that are associated with chlorophyll and oxidation products. The chlorophylls' bands (absorption between 300 and 650 nm, emission between 650 and 750 nm, R1 in Figure \ref{fig:results2}) are the most significant spectral components for the prediction of $K_{232}$. 
For the determination of $K_{268}$, the oxidation products (absorption between 300 and 400 nm, emission between 400 and 500 nm, R2 in Figure \ref{fig:results2}) acquire relevance. These latter results indicate that $K_{268}$ is the most sensitive indicator of the presence of oxidation products. This is consistent with previous observations of greater changes in $K_{268}$ than in $K_{232}$ during thermal degradation \cite{venturini2024shedding}.

\begin{figure}[hbt]
    \centering
    \includegraphics[width=1.0\textwidth]{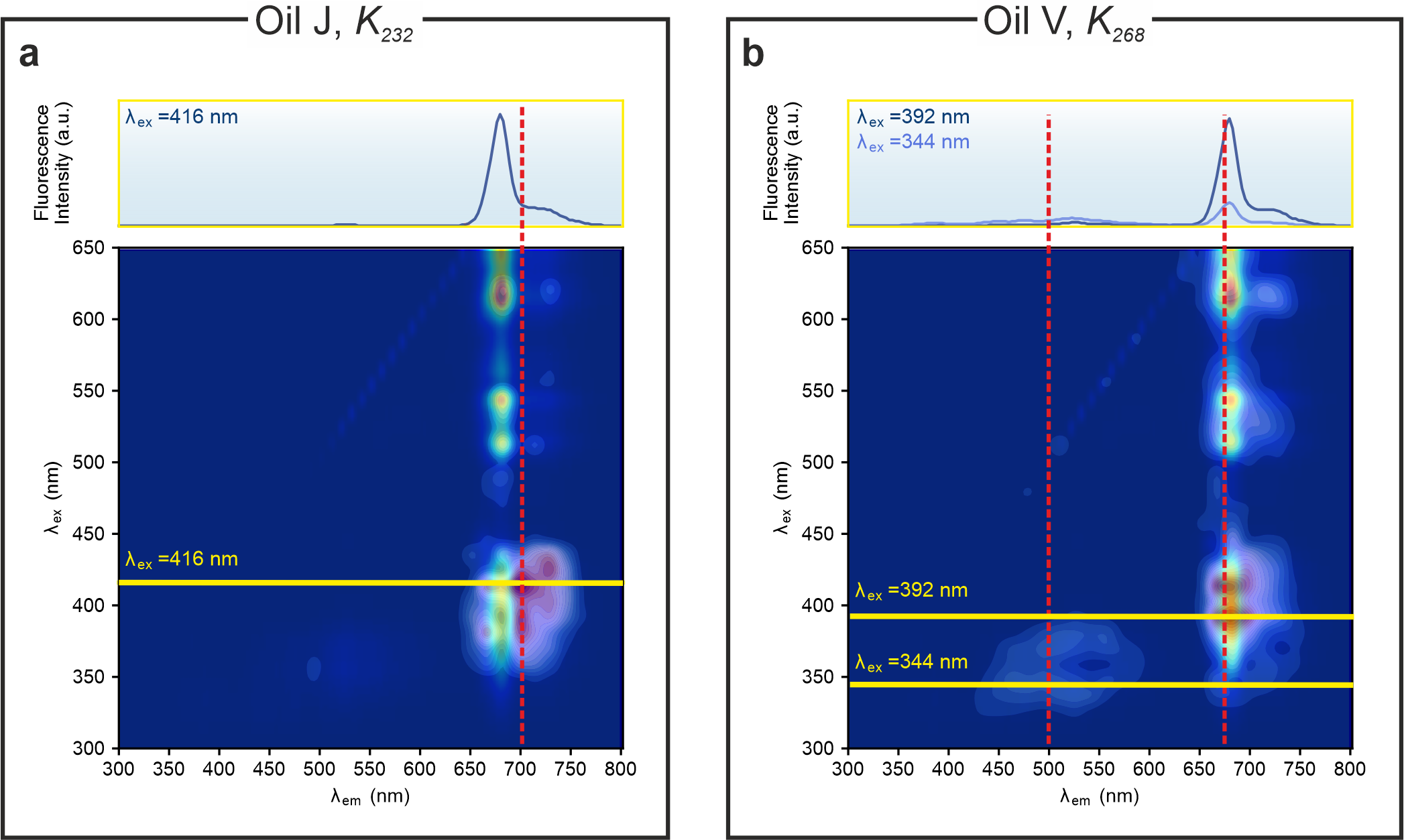}
    \caption{IEA approach showing the spectral bands of the fluorescence spectrum which most significantly contribute to the prediction for two selected oils in the last oxidation stage. \textbf{a}, EEM of Oil J and region importance heatmap overimposed for quality indicator $K_{232}$. The fluorescence spectrum at $\lambda_{\textrm{ex}}=416$ nm is shown in the top panel. \textbf{b}, EEM of oil V and region importance heatmap overimposed for quality indicator $K_{268}$. The  fluorescence spectra at $\lambda_{\textrm{ex}}=344$ nm and $\lambda_{\textrm{ex}}=392$ nm are shown in the top panel.}
    \label{fig:cuts}
\end{figure}

The analysis of the relevant spectra features can be further pushed by overlaying the heatmap with the EEMs, as shown in Figure \ref{fig:cuts}. The IEA specifically identifies as the most relevant excitation wavelength $\lambda_{\textrm{em}}=416$ nm. The most relevant features of the corresponding fluorescence spectrum for the prediction of $K_{232}$ are determined to be the shoulder between the maximum of the chlorophyll emission at 680 nm and the secondary broad peak at 720 nm. Differently, for the prediction of $K_{268}$ the IEA identifies as the most relevant excitation wavelengths $\lambda_{\textrm{em}}$ to be 344 nm and 392 nm. The most important feature of the corresponding fluorescence emission spectra is the maximum of the chlorophyll emission peak at 680 nm, with some contribution from oxidation products at ca. 500 nm. These findings show that the two quality indicators are affected by presence of different chemical components. 

\begin{figure}[hbt]
    \centering
    \includegraphics[width=1.0\textwidth]{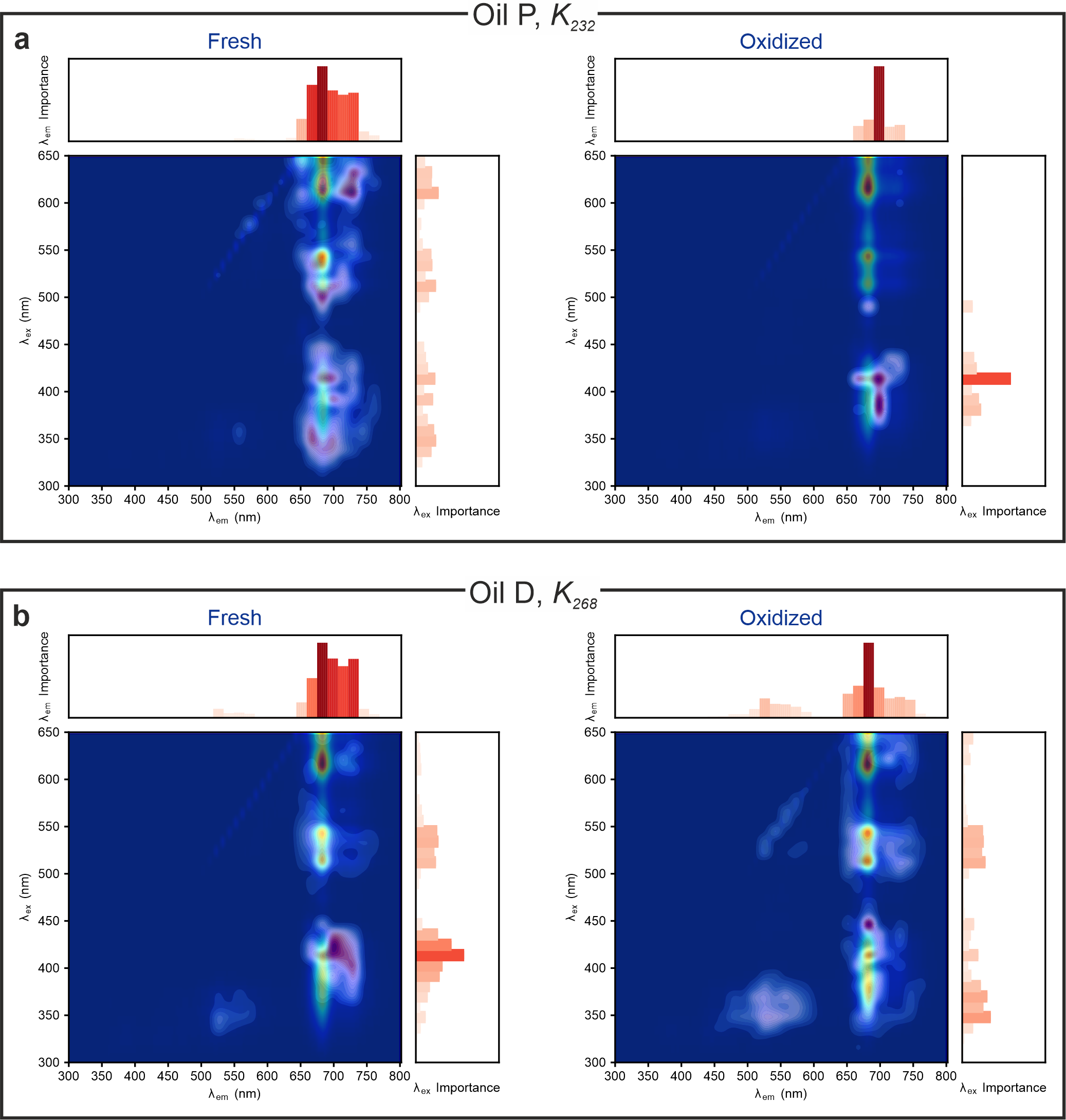}
    \caption{IEA approach showing the evolution during the oxidation process on two selected oils: ``Fresh'' marks the oil just after opening the bottle, ``Oxidized'' the oil at the latest oxidation stage. \textbf{a}, Oil P EEM and region importance heatmap overimposed for the prediction of the quality indicator $K_{232}$; \textbf{b}, Oil D and region importance heatmap overimposed for the prediction of the quality indicator $K_{232}$.}
    \label{fig:evolution}
\end{figure}

Finally, the IEA can be used to study the progression of the oxidation process by investigating the evolution of the identified spectral bands. The results of this analysis are shown in Figure \ref{fig:evolution} for two selected oils as an example.
As the oxidation progresses, the identified relevant spectral range becomes narrower, visually recognisable as a heatmap that is less spread throughout the EEM (see Figure \ref{fig:evolution}a) and if oxidation products are present, these acquire more relevance (see Figure \ref{fig:evolution}b). This means that the model becomes more precise in the identification of the relevant chemical components as the features emerge more clearly.

\section*{Conclusions}

In summary, the new contributions of this work are twofold. First, we present a domain adaptation method that allows the application of DL even with a small dataset. The method is demonstrated through its application to assessment of the decrease in quality of extra virgin olive oil during storage from fluorescence data. The model is trained on a dataset of 240 fluorescence EEM with high similarity to determine the mandatory indicators $K_{232}$ and $K_{232}$ needed for quality assessment. The results show that the model has an exceptional performance and predicts the two indicators $K_{232}$ and $K_{232}$ with a very high accuracy (low MAE), comparable to the experimental error. Therefore, the proposed approach allows to construct a model that is interpretable as a proxy for the physico-chemical process. This approach represents a significant innovation because it applies domain adaptation to utilize the MobileNetv2 architecture, which is originally designed and pretrained using conventional photographic images, for scientific applications, thus bridging the gap between classical image recognition and the specialised needs of scientific imagery analysis.
Second, we introduce the IEA which allows to determine without any prior knowledge the spectral absorption and emission bands involved in the oxidation process. The spectral bands are robustly identified independently of the specifics of the olive oil and become more and more pronounced as the process progresses, as expected. This second contribution transforms DL from a black-box to an interpretative tool, thus offering a key to understanding the underlying physico-chemical process. In general, the impact and applicability of the research presented in this work goes beyond fluorescence EEM because such an approach can be easily applied to other types of data.

%\section*{Disclosures}
%The authors declare no competing interests.

\section*{Data Availability}
The dataset of this study is freely available for download at \url{https://doi.org/10.17632/g6y69g8gwm.1} and is described in detail in \cite{venturini2023dataset}.

\bibliography{sn-bibliography}    % makes bibtex use spiejour.bst

\section*{Acknowledgments}
This acquisition of the dataset was supported by the Hasler Foundation project ``ARES: AI for fluoREscence Spectroscopy in oil''.

\section*{Author Contributions}
F.V. conceived the project, supervised the analysis, and wrote the paper. U.M. conceived the method, developed and implemented the code, and wrote the paper.

%%%%% Biographies of authors %%%%%
% \vspace{2ex}\noindent\textbf{Umberto Michelucci} is senior lecturer at the Lucerne University of Applied Sciences. He studied theoretical physics in Florence, Italy, focusing on theoretical simulations of atom entrapment with lasers and high-temperature superconductors. He holds a PhD in computer science and machine learning from the University of Portsmouth, with a focus in multi-task learning neural networks for optics and sensor science. He authored five books and several journal and conference papers. He is co-founder of TOELT LLC, a company focused on research in machine learning and he is a Google Developer Expert in Machine Learning.

% \vspace{1ex}\noindent\textbf{Francesca Venturini} is full professor at the Zurich University of Applied Sciences, Switzerland. She received her MS degrees in physics from the University of Florence, Italy in 1997, and her PhD degree from the Technical University of Munich, Germany in 2003. She is co-founder of TOELT LLC. She is the author more than 60 journal and conference papers and inventor in nine patents. Her research interests are optical spectroscopy and the development of new machine learning approaches for data analysis and the development of explainability techniques.

%\listoffigures

\end{spacing}
\end{document}